\documentclass[pdflatex,sn-vancouver-ay]{sn-jnl}
\usepackage{graphicx}%
\usepackage{multirow}%
\usepackage{amsmath,amssymb,amsfonts}%
\usepackage{amsthm}%
\usepackage{mathrsfs}%
\usepackage[title]{appendix}%
\usepackage{xcolor}%
\usepackage{textcomp}%
\usepackage{manyfoot}%
\usepackage{booktabs}%
\usepackage{algorithm}%
\usepackage{algorithmicx}%
\usepackage{algpseudocode}%
\usepackage{listings}%

\usepackage{fontspec}
\usepackage{polyglossia}
\setdefaultlanguage{english}
\setotherlanguage{bengali}

\newfontfamily\banglafont[Script=Bengali,Language=Bengali]{kalpurush.ttf}

\newcommand{\bn}[1]{{\banglafont #1}}

\begin{document}

\title[Distinguishing Repetition Disfluency from Morphological Reduplication in Bangla ASR Transcripts: A Novel Corpus and Benchmarking Analysis]{Distinguishing Repetition Disfluency from Morphological Reduplication in Bangla ASR Transcripts: A Novel Corpus and Benchmarking Analysis}

\author[1]{\fnm{Zaara Zabeen} \sur{Arpa}}\email{zaarazabeen@iut-dhaka.edu}

\author[1]{\fnm{Sadnam Sakib} \sur{Apurbo}}\email{sadnamsakib@iut-dhaka.edu}

\author[1]{\fnm{Nazia Karim Khan} \sur{Oishee}}\email{naziakarim@iut-dhaka.edu}

\author*[1]{\fnm{Ajwad} \sur{Abrar}}\email{ajwadabrar@iut-dhaka.edu}

\affil[1]{\orgdiv{Department of Computer Science and Engineering}, \orgname{Islamic University of Technology}, \orgaddress{\street{Board Bazar}, \city{Gazipur}, \postcode{1704}, \state{Dhaka}, \country{Bangladesh}}}

\abstract{Automatic Speech Recognition (ASR) transcripts, especially in low-resource languages like Bangla, contain a critical ambiguity: word-word repetitions can be either Repetition Disfluency (unintentional ASR error/hesitation) or Morphological Reduplication (a deliberate grammatical construct). Standard disfluency correction fails by erroneously deleting valid linguistic information. To solve this, we introduce the first publicly available, 20,000-row Bangla corpus, manually annotated to explicitly distinguish between these two phenomena in noisy ASR transcripts. We benchmark this novel resource using two paradigms: state-of-the-art multilingual Large Language Models (LLMs) and task-specific fine-tuning of encoder models. LLMs achieve competitive performance (up to 82.68\% accuracy) with few-shot prompting. However, fine-tuning proves superior, with the language-specific BanglaBERT model achieving the highest accuracy of 84.78\% and an F1 score of 0.677. This establishes a strong, linguistically-informed baseline and provides essential data for developing sophisticated, semantic-preserving text normalization systems for Bangla.}

\maketitle

\section{Introduction}
\begin{figure}[t]
    \centering
    \includegraphics[width=\columnwidth]{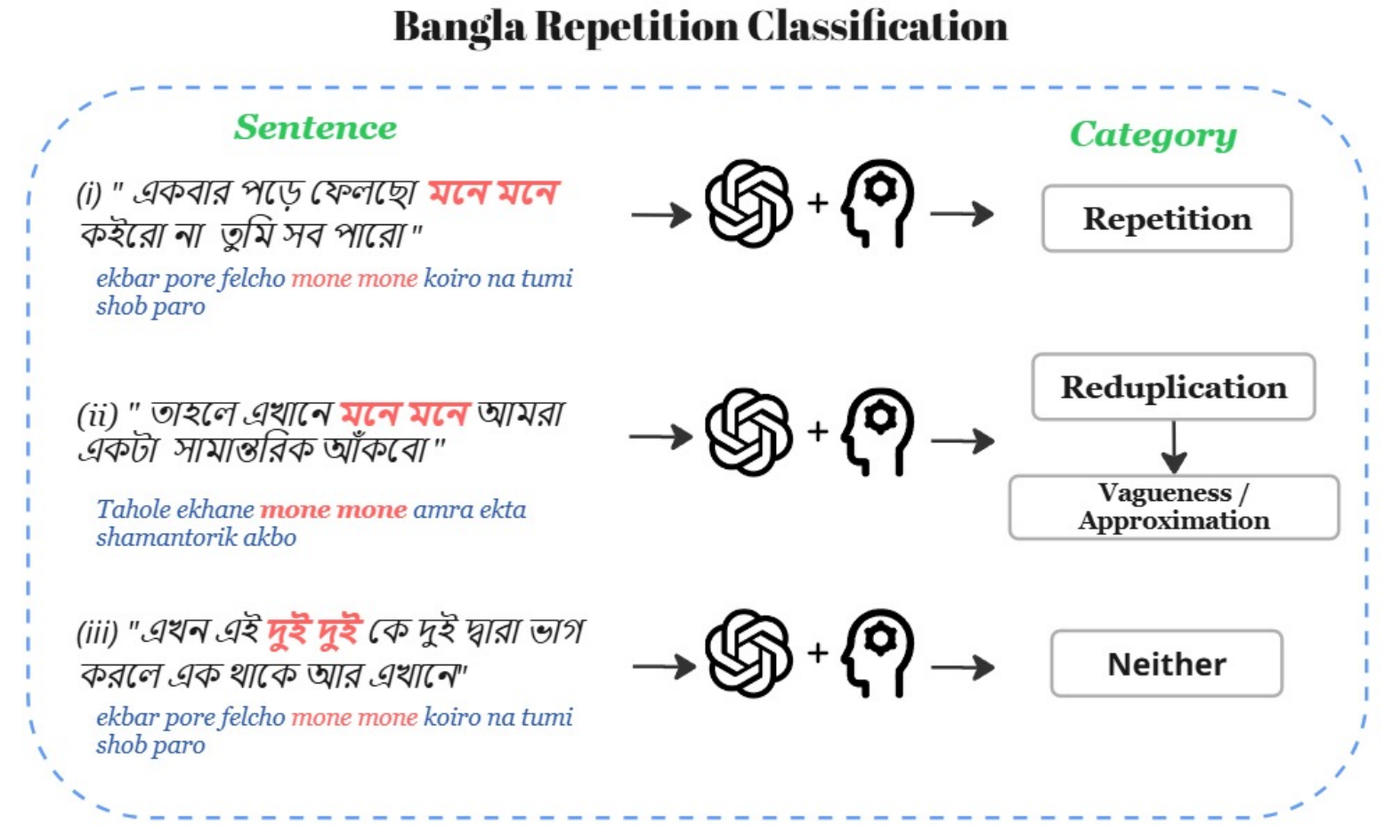}
    \caption{Illustration of the Bangla Repetition Classification task, highlighting the distinction between unintentional disfluencies (Repetition), grammatical forms (Reduplication), and coincidental occurrences (Neither).}
    \label{fig:Example of classification}
\end{figure}
\subsection{ASR Pitfalls: Disfluency and Repetition}
Automatic Speech Recognition (ASR) is now integral to digital interaction, driving applications from virtual assistants to automated subtitles on platforms like YouTube~\citep{uhrig2023open}. Despite its wide adoption and significant advancements, ASR performance remains imperfect, particularly in Large Vocabulary Continuous Speech Recognition (LVCSR) and under real-world conditions like background noise or diverse accents~\citep{errattahi2018automatic, riad2022automatic}.

This often results in a significant Word Error Rate (WER). A major, persistent source of error stems from speech disfluencies, which are interruptions in the smooth flow of speech, including filled pauses, hesitations, self-corrections, and most relevantly, repetitions~\citep{riad2022automatic}. Disfluencies are natural and frequent; one study found a $\mathbf{50\%}$ probability of a disfluency in a 10-13 word sentence~\citep{lou2020improving}. Their presence creates noisy transcripts that are difficult to read and detrimental to downstream Natural Language Processing (NLP) tasks such as machine translation or information extraction~\citep{riad2022automatic, jamshid-lou-johnson-2020-end}. Consequently, automatic Disfluency Correction (DC) is a critical research area, aiming to ``clean" ASR outputs~\citep{errattahi2018automatic}. High-quality DC corpora, such as DISCO, have enabled benchmark F1 scores up to $94.29\%$ in languages like Hindi, confirming DC's vital post-processing role~\citep{bhat-etal-2023-disco}.

\subsection{Bangla Ambiguity: Disfluency vs. Reduplication}
The conventional DC approach treats repetitions as universal ``noise" to be removed. This fails in languages with specific morphological properties that structurally align with disfluent phenomena. This paper addresses a critical ambiguity in Bangla, an Indo-Aryan language with over 270 million speakers~\citep{ridoy2025adaptability}. In Bangla ASR transcripts, the identical surface form $word$-$word$ can represent two phenomena with opposing grammatical implications:

\begin{enumerate}
    \item \textbf{Repetition Disfluency:} An unintentional, non-grammatical repetition (speaker hesitation or ASR error), often described by the Reparandum-Interregnum-Repair (RiR) framework~\citep{bhat-etal-2023-disco, ahmad-etal-2025-looks}. \emph{Example:} \bn{“এই যে পাঁচ তারিখ থেকে শুরু করে \textbf{করে} তোমার 12 তারিখ পর্যন্ত।”} (\emph{...shuru kore kore...}). Here, the repeated \bn{“করে”} (\emph{kore}) is an error and should be deleted.
    \item \textbf{Morphological Reduplication:} A deliberate, rule-governed, and grammatically significant process where a word is repeated to convey a specific semantic nuance, such as continuity, iterativity, intensity, or plurality~\citep{rana2010reduplication, abbi1992reduplication}. \emph{Example:} \bn{“অংকগুলো \textbf{করে করে} আমরা একটু আন্ডারস্ট্যান্ডিং ডেভেলপ করা চেষ্টা করবো”} (\emph{Onkogulo kore kore...}). The repeated phrase \bn{“করে করে”} (\emph{kore kore}) is a crucial construction conveying an iterative nature and must be preserved.
\end{enumerate}

This structural ambiguity, visually detailed in Figure~\ref{fig:Example of classification}, presents a formidable challenge. Generic disfluency detection models designed with a ``subtractive" philosophy would fail by erroneously stripping away valid linguistic information, catastrophically altering the semantic content of the text. Resolving this requires a fine-grained, context-aware classification model.

\subsection{The Low-Resource Data Gap}
Developing models to resolve such language-specific ambiguities requires large-scale, high-quality annotated data. Despite its massive speaker base, Bangla is a low-resource language in NLP~\citep{ridoy2025adaptability} due to a scarcity of standardized, publicly available datasets. For the specific task of distinguishing repetition disfluency from morphological reduplication in Bangla, no publicly available annotated corpus existed prior to this work. This resource gap has been the primary impediment to developing and rigorously evaluating computational systems for this task, hindering the shift from generic, one-size-fits-all NLP solutions toward models sensitive to the unique grammatical structures of low-resource languages.

\subsection{Contributions}
This paper addresses this critical resource and research gap by providing the necessary data and establishing strong performance baselines. The primary contributions are threefold:

\begin{enumerate}
\item \textbf{Corpus Creation:} We introduce the first publicly available, $\mathbf{20,000}$-row Bangla corpus, manually annotated to explicitly distinguish between Repetition Disfluency and Morphological Reduplication in noisy ASR transcripts. Furthermore, we provide a fine-grained linguistic analysis by subcategorizing all Morphological Reduplication instances into nine distinct semantic and functional classes.
\item \textbf{LLM Benchmarking:} We benchmark state-of-the-art multilingual Large Language Models (LLMs) (GPT, Gemini, Claude families) under zero-shot, one-shot, and few-shot prompting. LLMs achieve a competitive performance up to $\mathbf{82.68\%}$ accuracy with few-shot prompting.
\item \textbf{Fine-Tuning Analysis:} We empirically demonstrate the superiority of task-specific fine-tuning. The language-specific BanglaBERT~\citep{bhattacharjee-etal-2022-banglabert} model achieves the highest performance with an accuracy of $\mathbf{84.78\%}$ and an F1 score of $\mathbf{0.677}$, establishing a strong, linguistically-informed baseline for developing semantic-preserving text normalization systems for Bangla.
\end{enumerate}

\section{Related Works}
Our research is situated at the intersection of speech processing, computational linguistics, and low-resource NLP. We contextualize our contribution by reviewing the distinct treatment of repetition as an error in disfluency correction versus a meaningful construct in morphological reduplication, and by outlining the standard methodological paradigms our work builds upon.

\subsection{The Dichotomy of Repetition: Disfluency vs. Reduplication}
Computational Disfluency Correction (DC) is a critical post-processing step for ASR, designed to improve transcript readability by identifying and removing phenomena like filled pauses, self-corrections, and repetitions~\citep{riad2022automatic}. The field has evolved from classic sequence tagging to sophisticated Transformer-based architectures, with large-scale corpora like DISCO enabling high F1 scores in high-resource languages~\citep{bhat-etal-2023-disco, lou2020improving}. For low-resource languages, including Bengali, the lack of labeled data has spurred techniques like zero-shot learning with multilingual encoders and synthetic data augmentation via adversarial training to improve performance~\citep{kundu2022zero, bhat-etal-2023-adversarial, wang-etal-2022-adaptive}.

However, a fundamental limitation of the dominant DC paradigm is its inherently subtractive nature that treats all repetitions as ``noise" to be deleted~\citep{lou2020improving, jamshid-lou-johnson-2020-end}. This approach is incompatible with languages like Bangla, where repetition is also a productive grammatical device. In linguistics, morphological reduplication is a rule-governed process where a word is repeated to encode specific semantic nuances, such as continuity, iterativity, or intensity~\citep{rana2010reduplication, abbi1992reduplication}. This creates a critical structural ambiguity where the surface form `word-word` can be either an error or a meaningful linguistic construct.

This specific challenge is recognized across the Indo-Aryan language family. Recent parallel work has successfully benchmarked this classification task in Hindi, Marathi, and Telugu, achieving Macro F1 scores up to 85.62\% and demonstrating the necessity of context-aware models that can distinguish grammatical reduplication from disfluent structures like the Reparandum-Interregnum-Repair (RiR) pattern~\citep{ahmad-etal-2025-looks}. While early computational work on reduplication in Indic languages relied on rule-based systems or finite-state transducers~\citep{das2010identification, dolatian2019computational}, our work addresses this problem using modern neural architectures. We bridge the gap between the subtractive ASR-processing paradigm and principled linguistic analysis by creating a resource to train models for this nuanced classification task.

\subsection{Methodological Context and Modeling Paradigms}
Developing robust NLP solutions for Bangla is hindered by a scarcity of standardized datasets, a common challenge for low-resource languages despite their large speaker populations~\citep{ridoy2025adaptability}. This has motivated broad efforts to create foundational resources and models for the Indic language family~\citep{kakwani2020indicnlpsuite}. Our approach to corpus creation aligns with pragmatic solutions to this data gap: we leverage noisy, auto-generated ASR transcripts from YouTube~\citep{uhrig2023open}. This strategy is effective because the inherent flaws of ASR systems provide a naturalistic distribution of the very phenomena knowingly spurious repetitions, speaker hesitations, and correctly transcribed reduplications, required to train a robust real-world classifier.

For the classification task itself, we evaluate the two dominant paradigms for applying pre-trained models: in-context learning ~\citep{zhou-etal-2024-mystery} via prompting and task-specific fine-tuning~\citep{liu2023pre}. The former tests the ability of massive LLMs to perform the task with zero or few examples, while the latter adapts the weights of smaller, pre-trained encoder models to the specific dataset. Our experiments provide a direct comparison of these approaches, utilizing both general multilingual models (mBERT, XLM-RoBERTa) and the language-specific BanglaBERT~\citep{bhattacharjee-etal-2022-banglabert} to establish a strong, linguistically-informed baseline.
\begin{figure*}[t]
    \centering
    \includegraphics[width=\textwidth]{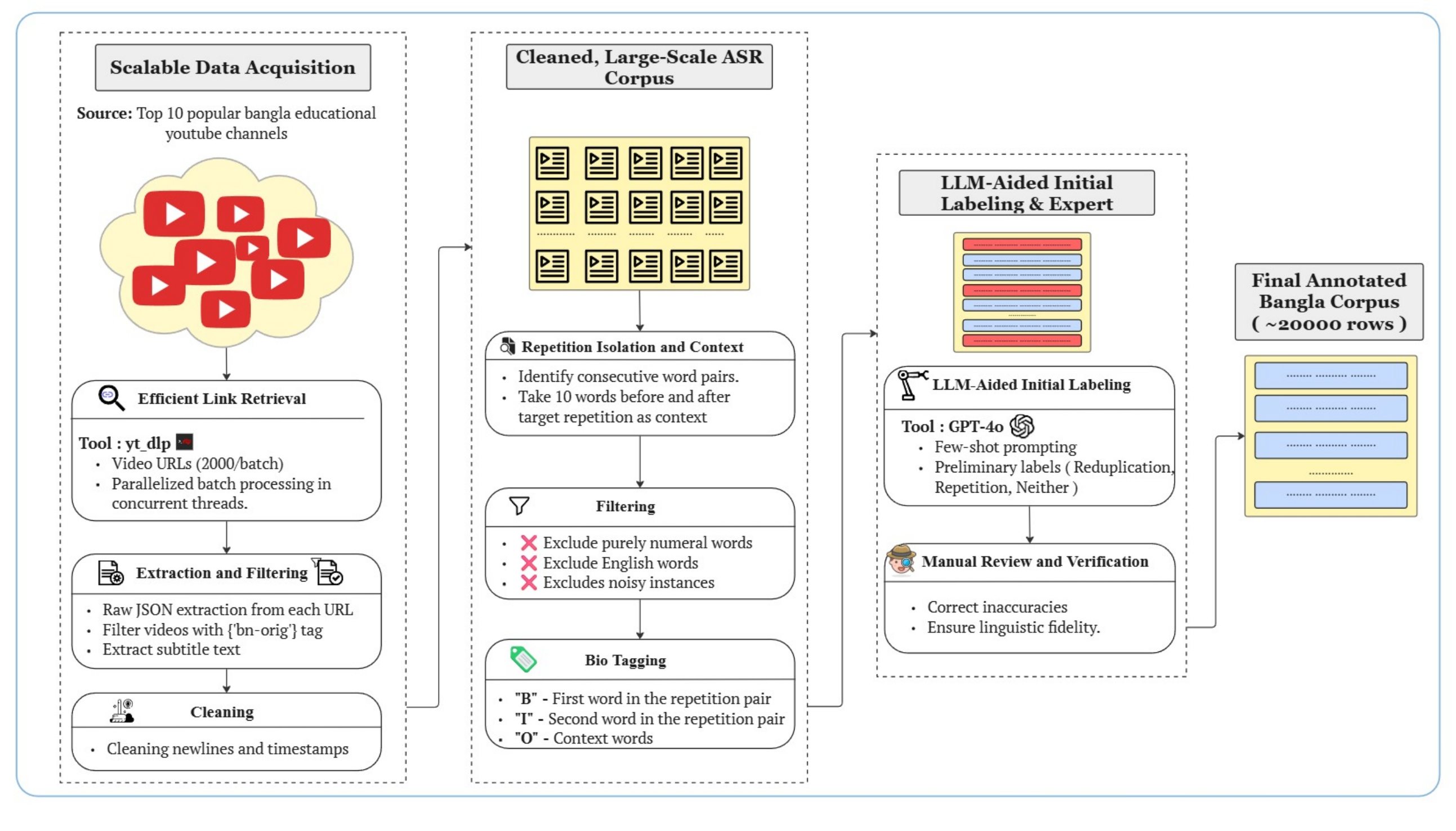}
    \caption{The end-to-end pipeline for creating the Bangla Repetition Corpus. The workflow begins with scalable data acquisition from YouTube ASR transcripts, followed by automated filtering and context extraction. The core annotation phase employs a hybrid approach, using an LLM for initial labeling and expert linguists for final verification, resulting in a 20,000-row gold-standard corpus.}
    \label{fig:methodology_overview}
\end{figure*}
\section{Methodology}
Our methodological framework is structured around three key phases: \textbf{Corpus Creation}, \textbf{LLM Benchmarking (Prompting)}, and \textbf{Task-Specific Fine-Tuning}. The comprehensive workflow for the corpus creation phase is illustrated in Figure~\ref{fig:methodology_overview}. This overall design establishes a strong performance baseline by rigorously comparing the capabilities of in-context learning against transfer learning for this fine-grained linguistic classification task.

\subsection{Corpus Creation: The Bangla Repetition Corpus}

The Bangla Repetition Corpus was synthesized from real-world, noisy Automatic Speech Recognition (ASR) transcripts, ensuring a naturalistic distribution of both errors and grammatical forms. This process involved four steps: Scalable Data Acquisition, Automated Filtering, Expert Annotation, and Fine-Grained Sub-categorization.

\subsubsection{Scalable Data Acquisition}
We selected the top 10 popular Bangla educational YouTube channels as the data source to ensure a high volume of continuous, spontaneous spoken Bengali.

\begin{enumerate}
    \item \textbf{Efficient Link Retrieval:} Video Uniform Resource Locators (URLs) were collected using the \texttt{yt-dlp} utility. To maximize throughput, the retrieval process was parallelized using concurrent threads. We processed content from the channels' ``videos" tabs in batches of $\mathbf{2000}$ to quickly compile a comprehensive list of video links.
    \item \textbf{ASR Transcript Filtering:} Only videos confirmed to contain Bangla ASR transcripts were retained. This was programmatically verified by checking for the presence of the \texttt{bn-orig} tag in the subtitle manifest, confirming the noisy nature of the source data.
    \item \textbf{Subtitle Extraction and Cleaning:} Subtitles were downloaded and extracted from the raw JSON format using a parallel process. The text segments were joined, and noise reduction included replacing newlines and standardizing whitespace.
\end{enumerate}

\subsubsection{Automated Filtering and Structural Tagging}
From the cleaned, large-scale corpus of ASR text, we automatically extracted the ambiguous cases of contiguous word repetition:
\begin{enumerate}
    \item \textbf{Repetition Isolation:} The corpus was processed in memory-efficient chunks. Consecutive, identical word pairs ($\text{word}_i = \text{word}_{i-1}$) were identified after tokenization using a regular expression that explicitly handles Bengali, English, and numerical tokens ($\texttt{[\u0980-\u09FF]+|[a-zA-Z0-9]+}$), while excluding purely numerical repetitions.
    \item \textbf{Context Window Formulation:} Each repetition instance was extracted alongside a fixed, symmetric context window of $\mathbf{10 \text{ preceding words and } 10 \text{ following words}}$. This step resulted in a set of approximately $\mathbf{29,000}$ sentences containing candidate repetition instances.
    \item \textbf{Initial BIO Tagging:} To facilitate manual annotation and set up the task for supervised models, the repeated words within the context window were pre-tagged using a BIO scheme: the first occurrence was marked $\mathbf{B}$ (Beginning) and the second was marked $\mathbf{I}$ (Inside/Continuation).
\end{enumerate}

\subsubsection{LLM-Aided Initial Labeling and Expert Annotation}
The large set of approximately $\mathbf{29,000}$ filtered sentences underwent a two-stage labeling process, combining the scalability of generative models with the precision of expert human review.

\begin{enumerate}
    \item \textbf{LLM-Aided Initial Labeling:} To accelerate the annotation of the vast corpus, the sentences were first processed using the leading commercial model, GPT-4o, as an initial categorization engine. The model utilized the structured, few-shot prompt strategy, providing preliminary labels (Reduplication, Repetition, or Neither) for the entire set.
    \item \textbf{Expert Manual Review and Verification:} The corpus with the preliminary LLM labels was then subjected to a rigorous manual review by expert Bengali speakers. This critical step served to correct any inaccuracies introduced by the LLM and to ensure linguistic fidelity, particularly for nuanced cases where the distinction between a complex reduplication form and a speaker hesitation was ambiguous.
    \item \textbf{Final Corpus Selection:} Following the comprehensive manual verification and cleaning, all sentences that could not be unambiguously classified (often due to extreme ASR noise or context fragmentation) were discarded. This process finalized the core corpus, resulting in a set of $\mathbf{20,000}$ gold-standard rows, which were subsequently used for training and evaluation.
\end{enumerate}

The final annotated $\mathbf{20,000}$ rows were categorized into the three mutually exclusive labels:

\begin{enumerate}
\item \textbf{Reduplication (Grammatical):} The repetition is an intentional, rule-governed morphological process that conveys semantic nuances such as iterativity, continuity, intensity, or plurality. \emph{Example:} \bn{“অংকগুলো \textbf{করে করে} আমরা একটু আন্ডারস্ট্যান্ডিং ডেভেলপ করা চেষ্টা করবো”} (\emph{Onkogulo $\mathbf{kore \text{ } kore}$ amra...})
\item \textbf{Repetition (Disfluency):} The repetition is an unintentional error, either a speaker-induced hesitation or an ASR transcription error. \emph{Example:} \bn{“এই যে পাঁচ তারিখ থেকে শুরু করে \textbf{করে} তোমার 12 তারিখ পর্যন্ত।”} (\emph{...shuru kore $\mathbf{kore}$ tomar...})
\item \textbf{Neither:} Coincidental repetitions that are not clear disfluencies or productive reduplications.
\end{enumerate}
The final distribution of the annotated corpus, which shows a significant imbalance, is presented in Table \ref{tab:category_distribution}.

\begin{table}[t]
\centering
\begin{tabular}{lc}
\toprule
\textbf{Category} & \textbf{Percentage} \\
\midrule
Reduplication & 66.3\% \\
Repetition & 32.9\% \\
Neither & 0.8\% \\
\bottomrule
\end{tabular}
\caption{Category Distribution of the Annotated Bangla Corpus}
\label{tab:category_distribution}
\end{table}

\subsubsection{Fine-Grained Sub-categorization of Reduplication}
Following the primary classification, all instances identified as Morphological Reduplication underwent a second stage of fine-grained annotation to determine their specific semantic function. This was accomplished using a Large Language Model guided by a carefully constructed few-shot prompt that defined nine distinct subcategories:
\textbf{Intensity/Emphasis}, \textbf{Frequency/Iteration}, \textbf{Continuity/Ongoing Action}, \textbf{Plurality/Multiplicity}, \textbf{Distributive/Separateness}, \textbf{Vagueness/Approximation}, \textbf{Echo Word/Rhyming}, \textbf{Reciprocal/Correlative}, and \textbf{Onomatopoeia}.
For instance, in the sentence \bn{“...এক্স এর ভ্যালু পেয়ে গেছি কত কত বলো...”} (...we got the values of x, tell me what what...), the repeated word \bn{“কত কত”} (\textit{koto koto}) implies an iterative query for multiple values, leading to its classification as \textbf{Frequency / Iteration}. This two-tiered annotation process enriches the corpus, providing a detailed linguistic layer for future research.

\subsection{Experimental Setup and Baselines}
The evaluation was conducted across two distinct experimental setups on the held-out test set of $\mathbf{335}$ sentences, comparing the efficacy of in-context learning against transfer learning.

\subsubsection{LLM Benchmarking with Prompting Strategies}
We established a prompting baseline by evaluating seven state-of-the-art Large Language Models (LLMs) on this classification task. This set included leading proprietary models (GPT-4o, Claude 4, and Gemini 2.5 Flash) and several prominent open-source alternatives (Gemma 3, Mistral 7b instruct, Llama 3 8b Instruct, and Phi-4). The models were tested under Zero-shot, One-shot, and Few-shot ($\text{N} \leq 5$) conditions.

Our prompting strategy utilized a strictly controlled setup:
\begin{itemize}
    \item \textbf{Inference Control:} The temperature was set to $\mathbf{0.1}$ to favor deterministic and stable classification outputs.
    \item \textbf{Structured Prediction:} All prompts enforced a Structured JSON-in, JSON-out format, requiring the model to output a single, valid JSON object containing only the predicted category, which minimizes parsing errors and enforces a consistent response structure.
    \item \textbf{Explicit Context:} For few-shot tests, the prompt included explicit linguistic definitions and examples for the three target categories (Reduplication, Repetition, and Neither) to guide the LLMs' in-context learning capability.
\end{itemize}

\subsubsection{Task-Specific Fine-Tuning of Encoder Models}
We established robust performance baselines by conducting task-specific fine-tuning on three prominent Transformer-based encoder models: a Bangla-specific model and two high-performing multilingual models. The task was framed as a three-way sequence classification (sentence-level classification into Reduplication, Repetition, or Neither).

\begin{itemize}
    \item \textbf{Models:} We selected BanglaBERT (a language-specific model pre-trained on a vast Bengali corpus), XLM-RoBERTa (base), and mBERT (two widely-used multilingual models).
    \item \textbf{Training Parameters:} All models were fine-tuned for 3 epochs with a Batch Size of 16 and a Learning Rate of 2e-5. A Weight Decay of 0.01 was applied, and the Max Length was set to 128 tokens.
\end{itemize}
This approach directly compares the efficacy of using general multilingual pre-trained knowledge (mBERT/XLM-R) against specialized language pre-training (BanglaBERT) when adapting to a novel, linguistically-sensitive classification task on limited, noisy data.
\section{Results}

\subsection{LLM Benchmarking with Prompting Strategies}
Table \ref{tab:llm_accuracy} presents the accuracy of different multilingual LLMs across the three prompting strategies.

\begin{table*}[t]
\centering
\begin{tabular}{lccc}
\toprule
\textbf{Models} & \textbf{Zero-shot (\%)} & \textbf{One-shot (\%)} & \textbf{Few-shot (\%)} \\
\midrule
Claude 4 sonnet & 76.41 & 80.29 & \textbf{82.68} \\
GPT-4o & 78.50 & \textbf{80.50} & 82.10 \\
Gemini 2.5 Flash & \textbf{78.51} & 76.72 & 81.49 \\
Phi-4 & 61.19 & 60.30 & 62.39 \\
Gemma 3.4b & 63.88 & 61.10 & 46.20 \\
Llama 3 8b Instruct & 56.12 & 54.33 & 53.73 \\
Mistral 7b instruct & 43.88 & 66.27 & 62.09 \\
\bottomrule
\end{tabular}
\caption{Accuracy (\%) of Different Prompting Techniques on the Bangla Corpus. Bold values indicate the peak accuracy achieved by each model across the different prompting strategies.}
\label{tab:llm_accuracy}
\end{table*}

\subsubsection{Prompting Strategy Effectiveness}
The zero-shot performance of the leading LLMs (e.g., Gemini 2.5 Flash at 78.51\%) demonstrates that massive multilingual pre-training imparts a strong baseline capability to resolve word repetition ambiguity, likely leveraging latent knowledge across the Indo-European family. For the top models, few-shot prompting was the most effective method, consistently boosting accuracy to over 81\%. This confirms that explicit in-context examples are necessary to guide the LLMs toward the subtle grammatical cues that distinguish morphological reduplication from disfluency. Claude 4, for instance, achieved the highest LLM performance at 82.68\%. However, providing examples was inconsistent for the lower-tier models, sometimes degrading performance, which suggests that their internal representations are less robustly aligned with the task, and their ability to generalize from few-shot examples is limited.

\subsubsection{Model Ranking}
The consistently high performance of Claude 4, GPT-4o, and Gemini 2.5 Flash across all prompting configurations established them as the top-performing models (Table \ref{tab:llm_accuracy}). Conversely, the relatively low and inconsistent results from open-source models like Gemma 3.4b and Llama 3 8b Instruct underscored the challenge of generalizing complex linguistic rules in a low-resource setting without substantial specialized pre-training.
\begin{itemize}
\item \textbf{Top-Performing:} Claude 4, GPT-4o, Gemini 2.5 Flash.
\item \textbf{Mid-Tier:} Mistral 7b instruct, Phi-4.
\item \textbf{Low-Performing:} Gemma 3 4b, Llama 3 8b Instruct.
\end{itemize}

\subsection{Impact of Task-Specific Fine-Tuning}
\begin{figure}[htbp]
\centering
\includegraphics[width=\columnwidth]{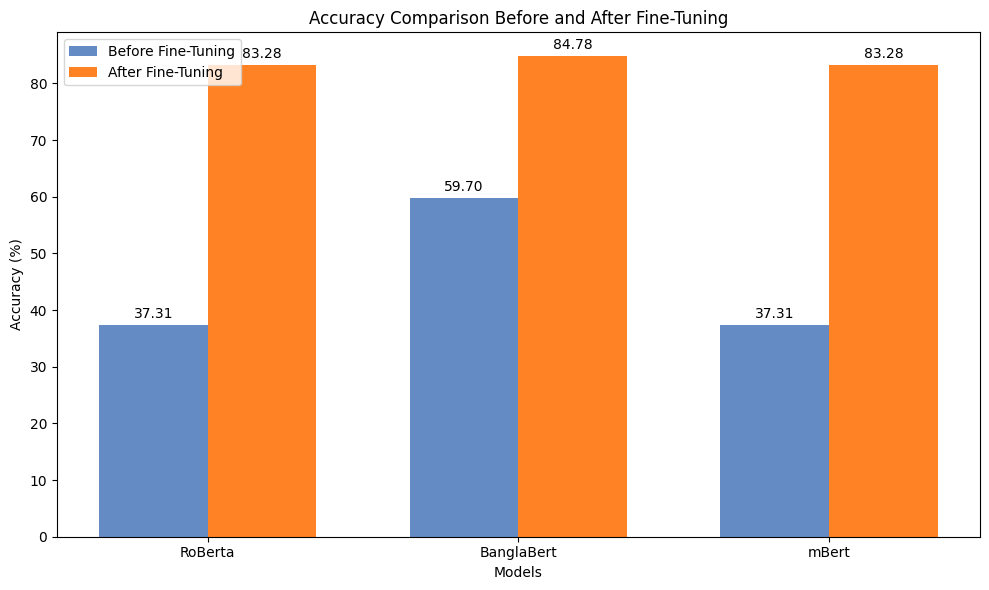}
\caption{Accuracy Comparison Before and After Fine-Tuning}
\label{fig:fine_tuning_acc}
\end{figure}

Fine-tuning the encoder models on our custom Bangla dataset led to substantial improvements in all metrics, significantly surpassing the highest performance achieved by the LLMs through prompting. Figure \ref{fig:fine_tuning_acc} illustrates the sharp gain in accuracy for all three models after fine-tuning.

The numerical results for accuracy and other key metrics are detailed in Table \ref{tab:fine_tuning_results}.

\begin{table*}[t]
\centering
\begin{tabular}{cccccc}
\toprule
\textbf{Model Name} & \textbf{Type} & \textbf{Accuracy (\%)} & \textbf{Precision} & \textbf{Recall} & \textbf{F1 Score} \\
\midrule
\multirow{2}{*}{\textbf{BanglaBERT}} & Base & 59.70 & 0.394 & 0.395 & 0.381 \\
 & Fine Tuned & \textbf{84.78} & \textbf{0.901} & \textbf{0.646} & \textbf{0.677} \\
\midrule
\multirow{2}{*}{\textbf{XLM-RoBERTa}} & Base & 37.31 & 0.267 & 0.331 & 0.190 \\
 & Fine Tuned & 83.28 & 0.556 & 0.580 & 0.566 \\
\midrule
\multirow{2}{*}{\textbf{mBert}} & Base & 37.31 & 0.124 & 0.333 & 0.181 \\
 & Fine Tuned & 83.28 & 0.553 & 0.581 & 0.565 \\
\bottomrule
\end{tabular}
\caption{Fine-tuning Results: Accuracy, Precision, Recall, and F1 Score. The fine-tuning process yields substantial gains in accuracy for all models (approx. 24-46 percentage points). \textbf{BanglaBERT} emerges as the strongest performer, achieving the highest Accuracy ($\mathbf{84.78\%}$) and a superior F1 Score ($\mathbf{0.677}$). The high Precision ($\mathbf{0.901}$) achieved by BanglaBERT is crucial, indicating a strong ability to preserve grammatically meaningful Reduplication instances, prioritizing semantic integrity over comprehensive disfluency removal.}
\label{tab:fine_tuning_results}
\end{table*}

\subsubsection{Key Fine-Tuning Findings}
\begin{itemize}
\item \textbf{BanglaBERT \citep{bhattacharjee-etal-2022-banglabert} Superiority and Cross-Linguistic Context:} BanglaBERT achieved the highest performance across the board after fine-tuning, with an accuracy of \textbf{84.78\%} and the highest precision ($\mathbf{0.901}$) and F1 score ($\mathbf{0.677}$). This result is competitive and consistent with state-of-the-art Macro F1 scores achieved in parallel research on this specific reduplication/repetition classification task in related Indo-Aryan languages, such as Hindi (up to 85.62\%) and Marathi (up to 84.82\%)~\citep{ahmad-etal-2025-looks}. This highlights the value of using a language-specific model for a nuanced task in a low-resource language.
\item \textbf{Fine-Tuning vs. Prompting:} The best fine-tuned model (BanglaBERT, $84.78\%$ accuracy) significantly outperformed the best-prompted LLM (Claude 4, $82.68\%$ accuracy), demonstrating that for this specific, linguistically-motivated classification task, the comprehensive parameter updates of fine-tuning are more effective than in-context learning.
\item \textbf{Metric Disparity Analysis:} A key observation is the substantial disparity between Precision ($\mathbf{0.901}$) and Recall ($\mathbf{0.646}$) for the Fine-Tuned BanglaBERT model. This disparity reflects the consequence of the highly imbalanced dataset (66.3\% Reduplication vs. 32.9\% Repetition). The high precision is highly desirable for normalization, as it indicates the model is extremely conservative, successfully avoiding False Positives (i.e., erroneously deleting meaningful Reduplication instances). Conversely, the lower recall shows that the model still misses a significant number of true Repetition Disfluency instances (False Negatives), allowing noise to remain in the transcript. This conservatism represents a strategic trade-off, prioritizing semantic preservation over comprehensive noise removal.
\item \textbf{Consistent Gains:} XLM-RoBERTa and mBERT both showed similar and substantial gains, reaching an accuracy of $83.28\%$. However, their F1 scores remained notably lower than BanglaBERT, reinforcing the advantage of specialized language pre-training.
\end{itemize}

\section{Conclusion}
This paper addresses the critical ambiguity between grammatical Morphological Reduplication and erroneous Repetition Disfluency in Bangla ASR transcripts. To solve this, we introduce the first publicly available, annotated corpus for this classification task. Our experiments demonstrate that task-specific fine-tuning is superior to few-shot prompting of large language models. The language-specific BanglaBERT model established the strongest performance baseline, achieving an accuracy of \textbf{84.78\%}. This work provides the essential data and a validated benchmark, paving the way for developing robust, semantic-preserving text normalization systems for Bangla.
\section*{Limitations and Future Work}
The primary limitation of this work stems from the high degree of dataset imbalance, with Reduplication instances significantly outnumbering Repetition instances (66.3\% vs. 32.9\%). While fine-tuning improved the F1 score, a substantial gap remains between precision and recall (e.g., BanglaBERT Fine Tuned: Precision 0.901, Recall 0.646), especially for the minority classes, suggesting that models may still be prone to bias towards the dominant Reduplication category. Future work must focus on mitigating this bias:

\begin{enumerate}
\item \textbf{Synthetic Data Augmentation:} We plan to leverage modern generative techniques to create a more balanced training environment. This includes using Large Language Models (LLMs) specifically as Disfluency Generators to create natural and diverse synthetic disfluent sentences for the minority class~\citep{cheng-etal-2024-boosting}. This strategy has been shown to be effective in capturing real-world disfluencies in low-resource settings~\citep{kundu2022zero}.
\item \textbf{Adversarial Training:} To improve the robustness of the fine-tuned model against noisy, real-world ASR outputs and enhance performance across all classes, we intend to implement Adversarial Training during the fine-tuning phase. This technique has previously yielded significant F1 improvements for Disfluency Correction tasks in Bengali and other Indian languages~\citep{bhat-etal-2023-adversarial}.
\end{enumerate}

Furthermore, the corpus is derived exclusively from educational content on YouTube. While this domain is rich in ASR errors and clear speech, it may not fully capture the linguistic variability, disfluency patterns, and reduplication nuances found in other spontaneous speech domains (e.g., political talk shows, casual vlogs, etc.), which could limit the generalizability of our model beyond this specific context. Future corpus expansion should target a more diverse range of conversational speech domains.

\begin{appendices}
\end{appendices}


\bibliography{sn-bibliography}

\end{document}